ORIGINAL PAPER

# Sense representations for Portuguese: experiments with sense embeddings and deep neural language models

**Jéssica Rodrigues da Silva**[1] · **Helena de M. Caseli**[1]

**Abstract** Sense representations have gone beyond word representations like Word2Vec, GloVe and FastText and achieved innovative performance on a wide range of natural language processing tasks. Although very useful in many applications, the traditional approaches for generating word embeddings have a strict drawback: they produce a single vector representation for a given word ignoring the fact that ambiguous words can assume different meanings. In this paper, we explore unsupervised sense representations which, different from traditional word embeddings, are able to induce different senses of a word by analyzing its contextual semantics in a text. The unsupervised sense representations investigated in this paper are: sense embeddings and deep neural language models. We present the first experiments carried out for generating sense embeddings for Portuguese. Our experiments show that the sense embedding model (Sense2vec) outperformed traditional word embeddings in syntactic and semantic analogies task, proving that the language resource generated here can improve the performance of NLP tasks in Portuguese. We also evaluated the performance of pre-trained deep neural language models (ELMo and BERT) in two transfer learning approaches: feature based and fine-tuning, in the semantic textual similarity task. Our experiments indicate that the fine tuned Multilingual and Portuguese BERT language models were able to achieve better accuracy than the ELMo model and baselines.

✉ Jéssica Rodrigues da Silva
jsc.rodrigues@gmail.com

Helena de M. Caseli
helenacaseli@ufscar.br

[1] Federal University of São Carlos (UFSCar), São Carlos, Brazil

## 1 Introduction

Any natural language (Portuguese, English, German, etc.) has ambiguities. This intrinsic feature causes the same word surface form to have two or more different meanings. Lexical ambiguities, which occur when a word has more than one possible meaning, directly impact tasks at the semantic level and solving them automatically is still a challenge in natural language processing (NLP) applications. One way to do this is through numeric vector representations.

**Word embeddings** are numerical vectors which can represent words or concepts in a low-dimensional continuous space, reducing the inherent sparsity of traditional vector-space representations (Salton et al. 1975). These vectors are able to capture useful syntactic and semantic information, such as regularities in natural language. They are based on the distributional hypothesis, which establishes that words that occur in the same contexts tend to have similar meanings (Harris 1954). The underlying idea that “a word is characterized by the company it keeps” was popularized by Firth (1957). A numerical vector representing a word can be visualized in a continuous vector space, accepting algebraic operations such as the cosine distance. This type of embedding is static, that is, it is generated only once during training and then used in NLP tasks.

The ability of word embeddings to capture knowledge has been exploited in several tasks, such as Machine Translation (Mikolov et al. 2013b), Sentiment Analysis (Socher et al. 2013), Word Sense Disambiguation (Chen et al. 2014) and Language Understanding (Mesnil et al. 2013).

Although very useful in many applications, the word embeddings, like those generated by Word2Vec (Mikolov et al. 2013a), GloVe (Pennington et al. 2014), Wang2Vec (Ling et al. 2015) and FastText (Bojanowski et al. 2017) have an important limitation: each word is associated with only one vector representation, ignoring the fact that polysemous words can assume multiple meanings. This limitation is called *Meaning Conflation Deficiency*, which is a mixture of possible meanings in a single word (Camacho-Collados and Pilehvar 2018). For instance, in the phrase “*My mouse was broken, so I bought a new one yesterday.*” the word “*mouse*” should be associated with its meaning of being a *computer device* rather than the *animal*. Figure 1 (Camacho-Collados and Pilehvar 2018) is an illustration of this Meaning Conflation Deficiency in a 2D semantic space.

In order to deal with the meaning conflation deficiency, several methods have been proposed to model individual word senses. The main distinction between them lies in how they model meaning. The first line of research that emerged aims at generating **sense embeddings** (Pina and Johansson 2014; Neelakantan et al. 2015; Wu and Giles 2015; Liu et al. 2015; Huang et al. 2012; Reisinger and Mooney 2010; Iacobacci et al. 2015), which model the meanings of a corpus following a knowledge-based or an unsupervised approach. After training, static sense embeddings are generated and can be used in NLP tasks.

Another more recent line of research has focused on the direct integration of embeddings into NLP tasks (Peters et al. 2018; Radford et al. 2018; Devlin et al. 2018; Liu et al. 2019a; Lample and Conneau 2019) by generating not a static, but a contextualized embedding, for which the representation changes dynamically depending on its occurrence context. These contextualized embeddings are also known as **deep neural language models**.

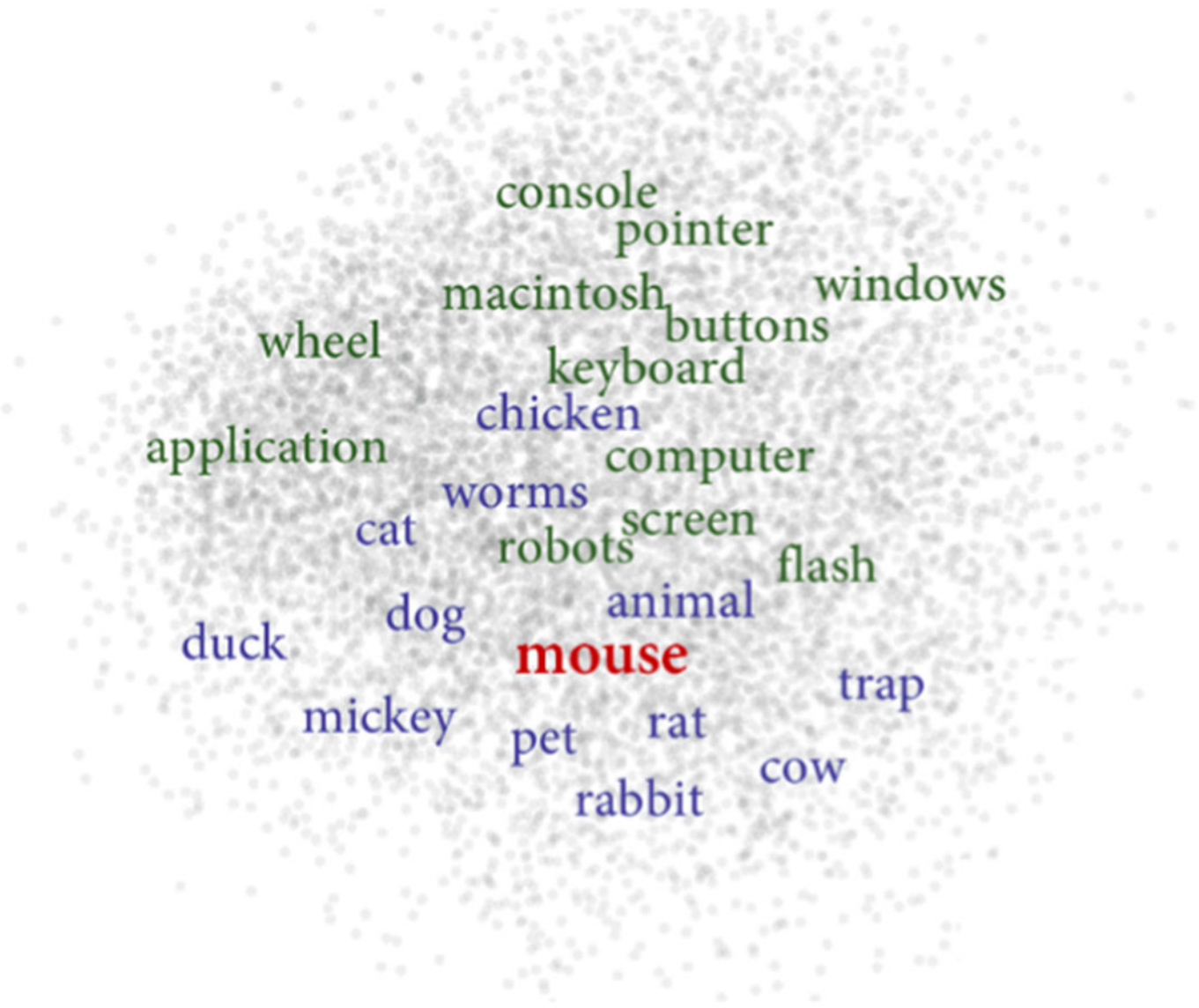

**Fig. 1** Example of meaning conflation deficiency of ambiguous word "mouse" from (Camacho-Collados and Pilehvar 2018). The words in blue refer to the sense of animal and the words in green to the sense of device

In this work, we present experiments using different types of sense representation for Portuguese. In a scenario where language resources are explored/generated on a large scale for English and Chinese, our work does this for Portuguese. Portuguese is a language with strong global relevance, with around 270 million speakers around the world, of which the largest community is in Brazil. It is a Romance language, so most of its lexicon is derived from Latin, but throughout its history it has incorporated many words from other languages.

According to Branco et al. (2012), Portuguese language has an inflectional paradigm which is much richer than that of languages such as English, particularly regarding verbs. For example, a verb can have different marks for aspect, time, mode, person, number, gender or polarity, reaching more than 160 different inflected forms, including simple and compound ones. Vector representations as word, sense and contextual embeddings are word-based embedding models that treat words as keys from which lower-dimensional embeddings can be looked up. What differentiates one word representation model from the others is the approach used to generate the vector and how they deal with context. Some methods take into account morphological information (as FastText), but the sense embeddings models investigated in this paper, MSSG and Sense2vec, are totally word-based, that is, they don't apply any technique that considers the morphology of the words for generating the vectors. Word-based language modeling relies on the distribution hypothesis (Harris, 1954) and most approaches to word vector representation use co-occurrence statistics. According to Takala (2016), these representations generalize poorly with morphologically rich languages, as vectors for all possible inflections cannot be generated (will not appear in the training corpus), and words with the same stem do not share a similar representation. This means that the quality of the resource is totally correlated with the language sparsity, which in our case is Portuguese. This means that the most similar words for the Portuguese verb "correr" (run, in English), for example, will not necessarily be the other forms of the verb such as "correndo", but similar semantic words such as "saltar" (jump, in English). However, for the English language, where the sparsity is less shocking, the three most similar words for the verb "run" are: "running", "runs" and "ran".[1]

Other challenges can be found when we take into account semantics. The scenario of lacking semantic resources pointed out by Branco et al. (2012), in 2012, is still a reality at the present moment although the number of available pre-trained resources is growing.[2] So, the development of sense embeddings for Portuguese and their evaluation in real NLP tasks are important steps towards evolving Portuguese resources for NLP.

Thus, the main contributions of this work are: (i) the generation and availability of pre-trained sense embeddings, (ii) the extensive experimentation of sense embeddings in intrinsic and extrinsic NLP tasks and (iii) the comparison of word and sense embeddings with the state-of-the-art deep neural language models. It is worth mentioning that all experiments covered the two variants of Portuguese: Brazilian and European. Finally, we also made available all the source code used in order to facilitate advances in this line of research.

This paper is organized as follows. In Sect. 2 we describe some of the approaches proposed in the literature. The methods investigated in this paper are described in Sect. 3. The experiments carried out with sense embeddings and language models for Portuguese and their evaluation are described in Sect. 4. Section 5 finishes this paper with some conclusions and proposals for future work.

## 2 Related work

Schütze (1998) was one of the first works to identify the meaning conflation deficiency of word embeddings and to propose the induction of meanings through the clustering of contexts in which an ambiguous word occurs. Then, many other works followed these ideas.

One of the first works using neural network to investigate the generation of sense embeddings was Reisinger and Mooney (2010). The method proposed by those authors is divided in two phases: (i) pre-processing and (ii) training. In the pre-processing phase, firstly, the **context** of each target word is defined as the words to the left and to the right of that target word. Then, each possible context is represented by the weighted average of the vectors of the words that compose it. These **context vectors** are grouped in clusters and each centroid is selected to represent the sense of the cluster. Finally, each word of the corpus is associated with the closest centroid to its context. After this pre-processing phase, a neural network is trained from the labeled corpus, generating the sense embeddings.

[1] These most similar words were obtained from the sense embeddings generated by MSSG for PT and EN based on wiki monolingual corpus (not parallel) with the same size. These sense embeddings are available with all the source code at the git of this project.

[2] Some pre-trained models for Portuguese can be found at http://www.davidsbatista.net/blog/2019/11/03/Portuguese-Embeddings/ and https://opencor.gitlab.io/corpora/souza20pre/.

In the experiments, Reisinger and Mooney (2010) used two training corpora: (1) a Wikipedia dump in English containing 2.8M articles with a total of 2.05B words and (2) the third English edition of Gigaword corpus, with 6.6M of articles and 3.9B words. The authors obtained a Spearman correlation of around 62.5% in WordSim-353 (Finkelstein et al. 2001),[3] for the Wikipedia and Gigaword corpus.

Huang et al. (2012) extends the Reisinger and Mooney (2010)'s method by incorporating a **global context** during the generation of sense embeddings. According to them, aggregating information from a larger context improves the quality of vector representations of ambiguous words that have more than one possible local context. To provide the vector representation of the global context, the proposed model uses all words in the document in which the target word occurs, incorporating this representation into the local context. The authors trained the model in a Wikipedia dump (from April 2010) in English with 2 million articles and 990 million tokens. The authors obtained a Spearman correlation of 65.7% in the Stanford's Contextual Word Similarities (SCWS),[4] surpassing the baselines.

Based on Huang et al. (2012), Neelakantan et al. (2015) proposed the generation of sense embeddings by performing an adaptation in the Skip-Gram model (Mikolov et al. 2013a) which they called Multiple-Sense Skip-Gram (**MSSG**). In MSSG, the identification of the senses occurs together with the training (one phase) to generate the vectors, making the process efficient and scalable. The authors used the same corpus as Huang et al. (2012) for training the sense embeddings and obtained a Spearman correlation of 67.3% also in the SCWS, surpassing the baselines. MSSG was one of the methods chosen to be investigated in this paper as described in Sect. 3.

Trask et al. (2015) proposes a different approach (**sense2vec**) that uses a tagged corpus rather than a raw corpus for sense embeddings generation. The authors annotated the corpus with part of speech (PoS) tags and that allowed the identification of ambiguous words from different classes. For example, this approach, if applied to Portuguese, would allow the distinction between the noun *livro* (book) and the verb *livro* (free), since depending on the context in which this word occur, it will receive one of these two labels. After PoS tagging the training corpus, they trained a word2vec (CBOW or Skip-Gram) model (Mikolov et al. 2013a). The authors did not report results comparing their approach with baselines. In addition to the PoS tags, the authors also tested the ability of the method to disambiguate named entities and sentiment, also labeling the corpus with these tags before generating sense embeddings. The sense2vec was one of the methods chosen to be investigated in this paper as described in Sect. 3.

[3] WordSim-353 is a dataset with 353 pairs of English words for which similarity scores were set by humans on a scale of 1 to 10.

[4] The SCWS is a dataset with 2003 word pairs in sentential contexts.

For Portuguese, Rodrigues and Branco (2018) present a model trained with word2vec in a corpus of 2.2 billion of tokens. The model was evaluated only in intrinsic tasks. The first one was the word analogy task (with the same test set used in our intrinsic experiment), where it is necessary to predict the fourth from three words. The second one was a lexical similarity task, which consists of calculating a semantic similarity score between two words. The third one was a conceptual categorization task, which consists of clustering a set of words into categories taking into account the semantic relations across those words. For the word analogy task, the best configuration of the model obtained 47.1% accuracy, reaching one of the best scores obtained in this task for Portuguese.

In contrast to the methods described so far, which directly learn sense representations from corpora, the method presented in Pelevina et al. (2017) induces a sense inventory from clustering of word embeddings, based on vector similarities. Experiments on two datasets, including a SemEval challenge on word sense induction and disambiguation, show that this method performed comparably to the state of the art.

All the approaches to generate sense embeddings presented so far are **unsupervised** ones, including the ones investigated in this work. However, there are several methods which rely on a knowledge base (KB) to provide sense information, the **knowledge-based** ones. According to Camacho-Collados and Pilehvar (2018), the knowledge-based methods represent word senses as defined by an external sense inventory (e.g., WordNet) while the unsupervised ones induce the sense distinctions by analyzing the occurrences of words in corpora.

In Bordes et al. (2011), a new neural network architecture was designed to embed any KB into a more flexible continuous vector space in which the original knowledge is kept and enhanced. The authors illustrate the method on WordNet and Freebase and also present a way to adapt it to knowledge extraction from raw text.

Another approach that uses sense inventories of knowledge bases was presented by Camacho-Collados et al. (2015). The proposed model was evaluated using two standard benchmarks of semantic similarity and Word Sense Disambiguation (WSD). In the semantic similarity task (word similarity), the proposed model outperformed the baselines (four other approaches that exploit Wikipedia as their main knowledge base) through the Pearson and Spearman correlations, in English, German and French datasets. For the WSD task, the authors chose the SemEval 2013 all-words that provides datasets for the Italian, English, French, Spanish and German languages (1123 words to disambiguate in each language). The model proposed by those authors outperformed the baselines—state-of-the-art WSD system and Most Frequent Sense (MFS)—through the F1 score for French, Spanish and German languages.

In what we can maybe can a hybrid method, Rothe and Schütze (2015) combined word embeddings with WordNet synsets to obtain sense embeddings. The approach was evaluated on lexical sample tasks by adding synset embeddings as features to an existing WSD system.

One advantage of knowledge-based methods is the use of sense inventories, making possible the generation of sense embeddings for each sense of this inventory. This facilitates the use and interpretation of the vectors, however, it also becomes a disadvantage because it makes impossible to discover new senses.[5]

In this work we chose to investigate unsupervised methods instead of inducing sense embeddings from any KB available for Portuguese. By doing this, we believe that the resources derived from this work can help to improve the NLP tasks for Portuguese without demanding a KB.

More recently, new proposals for language model generation—like ELMo (Peters et al. 2018), OpenAI GPT (Radford et al. 2018) and BERT (Devlin et al. 2018)—have begun to use more complex architectures to model context and capture the meanings of a word. The sequence tagger from Li and McCallum (2005) is one of the pioneer works that employ contextualized representations. The model infers context-sensitive variables for each word and integrates them with a CRF (Conditional Random Field) sequence tagger. Context2vec (Melamud et al. 2016) is one of the first and most prominent proposals in the new branch of **contextualized representations**. The model represents the context of a word extracting the output vector from a multilayer perceptron built on a bidirectional LSTM language model. Context2vec forms the basis for many of the subsequent works.

The Context Vectors (CoVe) model from McCann et al. (2017) calculates contextualized representations using a two layer bidirectional LSTM network in a sequence-by-sequence machine translation (seq2seq) model with attention. To evaluate their model, the authors performed a wide variety of common NLP tasks (for English): sentiment analysis, question classification, entailment, and question answering. For fine-grained sentiment analysis and entailment, CoVe improved the performance of baseline models to the state-of-the-art.

The ELMo (Embeddings from Language Model) uses a multi-layer LSTM language model in monolingual texts and some weights are shared between the two directions of the model. ELMo outperformed baselines across six benchmark NLP tasks (for English): question answering, textual entailment, semantic role labeling, coreference resolution, named entity extraction and sentiment analysis.

Following the idea of ELMo, OpenAI GPT (Generative Pre-training Transformer) expands the unsupervised language model to a much larger scale, training in a huge collection of raw text. It differs from ELMo in architecture and in the use of contextualized representations in NLP tasks. The authors performed experiments on a variety of supervised tasks (for English) including natural language inference, question answering, semantic similarity and text classification. For semantic similarity task, through Pearson correlation and F1 score, the model obtained state-of-the-art results on STS-B (Semantic Textual Similarity benchmark) and QQP (Quora Question Pairs dataset) with a 1 point absolute gain on STS-B.

BERT (Bidirectional Encoder Representations from Transformers) is a direct descendant of GPT: a large language model trained in raw text and tuned for specific NLP tasks, without custom network architectures. Compared to GPT, the biggest difference and improvement of BERT is to perform bidirectional training. The BERT model obtains new state-of-the-art results on eleven natural language processing tasks (for English), including pushing the GLUE score to 80.5% (7.7% point absolute improvement). BERT obtains a Pearson correlation score of 86.5 on the STS-B against 80.0 of the OpenAI GPT. For the QQP dataset, BERT obtains an F1 score of 72.1 against 70.3 of the GPT.

[5] Some KBs available for Portuguese are the OpenWordNet-PT (https://github.com/own-pt/openWordnet-PT), the WordNet.PT (http://wordnet.pt/) and the Onto.PT (http://ontopt.dei.uc.pt/).

A limitation of this approach to contextualized representation is the high computational cost due to the size/complexity of the models and the corpus. In this paper we have carried out experiments with ELMo and BERT, as described in Sect. 4.

## 3 Unsupervised sense representations

Unsupervised sense representations are built based only on the information extracted from text corpora. An unsupervised model induces different senses of a word by analyzing its contextual semantics in a text corpus and represents each sense based on the statistical knowledge derived from the corpus. We used two types of unsupervised sense representations: sense embeddings and contextualized embeddings. Each one is explained in the next sections.

### 3.1 Sense embeddings

Sense embeddings are one of the solutions to deal with the meaning conflation deficiency of word embeddings. The idea is to represent each meaning of a word (word sense) as a different vector.

In this paper, two unsupervised approaches were used for sense embeddings generation: the MSSG (Neelakantan et al. 2015) and the Sense2Vec (Trask et al. 2015).

#### *3.1.1 Multiple-Sense Skip-Gram (MSSG)*

In Neelakantan et al. (2015), two methods were proposed for generating sense embeddings based on the original Skip-Gram model (Word2vec) (Mikolov et al. 2013a): MSSG (Multiple-Sense Skip-Gram) and NP-MSSG (Non-Parametric Multiple-Sense Skip-Gram). The main difference between them is that MSSG implements a fixed amount of possible meanings for each word while NP-MSSG does this as part of its learning process. In both methods, the vector of the context is given by the weighted average of the vectors of the words that compose it. The context vectors are clustered and each cluster represents a sense existing in the corpus. Each cluster (sense) is associated to the words of the corpus by similarity to their context. After predicting the sense, the gradient update is performed on the centroid of the cluster and the training continues. The training stops when each word in the vocabulary has a vector representation generated for it. Different from the original skip-gram, MSSG and NP-MSSG learn multiple vectors for a given word. They were based on works such as Huang et al. (2012) and Reisinger and Mooney (2010). In this paper, we choose to work with the MSSG fixing the amount of senses for each target word. We did that to allow a fair comparison with the second approach (sense2vec) investigated here which works with a limited amount of meanings.

#### *3.1.2 Sense2Vec*

Trask et al. (2015) proposes the generation of sense embeddings from a corpus annotated with part-of-speech (PoS) tags, making it possible to identify ambiguous words from the amount of PoS tags they receive. For example, in Portuguese, the word *banco* (bank) would receive the noun tag if the context was about a financial institution and would receive the verb tag if the context was about banking / paying for something. The authors suggested that annotating the corpus with PoS tags is a costless approach to identify the different context of ambiguous words from different grammatical classes since these words will have a different PoS tag in each context. This approach makes it possible to create a meaningful representation for each use. The final step is to train a word2vec model (CBOW or Skip-Gram) (Mikolov et al. 2013a) with the tagged corpus, so that instead of predicting a word given neighboring words, it predicts a sense given the surrounding senses.

### 3.2 Contextualized embeddings

Unsupervised methods often generate sense embeddings through context clustering. As a consequence, the induced meanings may not be so clear nor easy to map to well-defined concepts. Recently, an emerging line of research has focused on the direct integration of vector representations into NLP tasks by means of **contextualized embeddings**.

Contextualized representations are context sensitive, that is, their representation changes dynamically depending on the context in which the target word appears. ELMo, BERT, and GPT-2 are architectures that generate contextualized embeddings (also known as deep neural language models) that are fine-tuned to create models for a wide variety of downstream NLP tasks. In these language models, the internal representations of words are called contextualized word representations because they are a function of the entire input sentence, and sentence embeddings are built through the summation of these representations. The success of this approach suggests that these representations capture highly transferable and task-agnostic properties of natural languages (Liu et al. 2019a). We have two transfer learning approaches to follow when working with neural language models: feature based or fine-tuning. According to Peters et al. (2019), in feature extraction the model's weights are 'frozen' and the pre-trained representations are used in a downstream model similar to classic feature-based approaches (Word2vec, GloVe, Sense2vec). Alternatively, pre-trained model's parameters can be unfrozen and fine-tuned on a new task. We choose both approaches in order to understand the performance of each one and the advantages of fine-tuning. In this paper, two pre-trained deep neural language models were used: ELMo and BERT .[6]

[6] GPT model was not used in our experiments since it was not yet available for the Portuguese language at the moment of the submission of this paper.

### *3.2.1 ELMo*

ELMo stands for Embeddings from Language Models and, hence, it has the fundamental ability of language models: to predict the next word in a sentence. When trained on a large dataset, the model also starts to pick up on language patterns. It consists of a two-layer bidirectional LSTM language model, built over a context independent character CNN layer and originally trained on the Billion Word Benchmark dataset (Chelba et al. 2013), consisting primarily of newswire text. In order to obtain a representation for each word, we performed a linear concatenation of all three ELMo layers, without learning any task-specific weights. For Portuguese, we used the model available through the AllenNLP library.[7] We intend to train ELMo from scratch with our corpus in future work.

### *3.2.2 BERT*

BERT is a deep Transformer (Vaswani et al. 2017) encoder trained jointly as a masked language model and on next-sentence prediction, originally trained on the concatenation of the Toronto Books Corpus (Zhu et al. 2015) and the English Wikipedia. In the experiments presented in this paper, we used two pre-trained BERT models: (i) the publicly released BERT-Base-multilingual model[8] and (ii) the BERTimbau-Base,[9] a Portuguese BERT model. The versions of the models used are available in the Transformers.[10] library. The BERT-multilingual model was simultaneously trained on the Wikipedia dumps for 104 different languages, which includes the Portuguese language. The Portuguese BERT was trained on the BrWaC (Brazilian Web as Corpus), a large Portuguese corpus with 2.7 billion tokens (Wagner Filho et al. 2018). With BERT, we decided to explore both ways of using deep neural language models: feature-based and the fine-tuning approaches. In order to achieve better accuracy on the semantic similarity task, we considered only the final layer of the models for generating the sentence embeddings.

We chose to use a pre-trained models due to the high computational cost to train one from scratch.

## 4 Experiments and results

In this section, experiments with sense embeddings and deep neural language models are presented. For sense embeddings, we trained the Sense2vec and MSSG models from scratch, in our Portuguese corpus. For the ELMo and BERT language models, we used pre-trained models.

7 https://allennlp.org/elmo.

8 https://github.com/google-research/bert/blob/master/multilingual.md.

9 https://github.com/neuralmind-ai/portuguese-bert/.

10 https://huggingface.co/transformers/.

**Table 1** Statistics of our training corpus

| Corpus | Tokens | Types | Genre |
| --- | --- | --- | --- |
| LX-Corpus (Rodrigues et al. 2016) | 714,286,638 | 2,605,393 | Mixed genres |
| Wikipedia | 219,293,003 | 1,758,191 | Encyclopedic |
| GoogleNews | 160,396,456 | 664,320 | Informative |
| SubIMDB-PT | 129,975,149 | 500,302 | Spoken language |
| G1 | 105,341,070 | 392,635 | Informative |
| PLN-Br (Bruckschen et al. 2008) | 31,196,395 | 259,762 | Informative |
| Literacy works ofpublic domain | 23,750,521 | 381,697 | Prose |
| Lacio-web (Aluísio et al. 2003) | 8,962,718 | 196,077 | Mixed genres |
| Portuguese e-books | 1,299,008 | 66,706 | Prose |
| Mundo Estranho | 1,047,108 | 55,000 | Informative |
| CHC | 941,032 | 36,522 | Informative |
| FAPESP | 499,008 | 31,746 | Science |
| Textbooks | 96,209 | 11,597 | Didactic |
| Folhinha | 73,575 | 9207 | Informative |
| NILC subcorpus | 32,868 | 4064 | Informative |
| Para Seu Filho Ler | 21,224 | 3942 | Informative |
| SARESP | 13,308 | 3293 | Didactic |
| Total | 1,395,926,282 | 3,827,725 | |

## 4.1 Training corpus

The corpus used for the training of sense embeddings was the same as Hartmann et al. (2017) which is composed of texts written in Brazilian Portuguese (PT-BR) and European Portuguese (PT-EU). In that work, a large corpus from several sources was collected in order to obtain a multi-genre corpus, representative of the Portuguese language, which was used to generate traditional word embeddings using the main well-know tools. Table 1 summarizes the information about these corpora: name, amount of tokens and types and a brief description of their genres.

Similar to what was done in Hartmann et al. (2017) for generating traditional word embeddings, the corpus was pre-processed in order to reduce the vocabulary size, under the premise that vocabulary reduction provides more representative vectors. Thus, numerals were normalized to zeros; URL's were mapped to a token URL and emails were mapped to a token EMAIL. Then, we tokenized the text relying on whitespaces and punctuation signs, paying special attention to hyphenation.

For the Sense2vec model, the corpus was PoS-tagged using the nlpnet tool (Fonseca and Rosa 2013), which is considered the state-of-art in PoS-tagging for PT-BR.

An example of a sentence with and without PoS tag is shown in Table 2. In this example, the ambiguous word *marca* (brand) occurs in sentences with two different

**Table 2** Example of sentences tagged with PoS in which the ambiguous word *marca* occurs

<table>
<tr><td>marcasense1</td><td>também virou modelo de uma marca famosa de roupas .</td></tr>
<tr><td></td><td>também|PDEN virou|V modelo|N de|PREP uma|ART marca|N famosa|ADJ de|PREP roupas|N .|PU</td></tr>
<tr><td>marcasense2</td><td>o relógio marca 00h, e o filme já vai começar .</td></tr>
<tr><td></td><td>o|ART relógio|N marca|V 00|NUM h|N ,|PU e|KC o|ART filme|N já|ADV vai|V começar|V .|PU</td></tr>
</table>

meanings: a manufacturer's trademark (sense1) and the infinitive of the verb *marcar* (mark) (sense2).

It is important to say that both approaches for generating sense embeddings were trained with this corpus. The only difference is that the input for the MSSG is the sentence without any PoS tag while the input for the sense2vec is the sentence annotated with PoS tags.

## 4.2 Network parameters

For all training, including baselines, we generated vectors of 300 dimensions, using the Skip-Gram model, with context window of five words to the left and five words to the right of the target word. The learning rate was set to 0.025 and the minimum frequency for each word was set to 10. For the MSSG approach, the maximum number of senses per word was set to 3.

## 4.3 Evaluation

The generated sense embeddings were intrinsically and extrinsically evaluated. For the intrinsic evaluation, the dataset of syntactic and semantic analogies of Rodrigues et al. (2016) was used. For the extrinsic evaluation, the sense embeddings were applied to tasks of Semantic Textual Similarity and Word Sense Disambiguation. The deep neural language models were evaluated only in the Semantic Textual Similarity task. Our baseline are word embeddings trained with Word2Vec, FastText and GloVe.

### *4.3.1 Intrinsic evaluation*

This experiment is a task of syntactic and semantic analogies where the use of sense embeddings is evaluated. Word embeddings were chosen as baselines.

*Dataset*. The dataset of Syntactic and Semantic Analogies of Rodrigues et al. (2016) has analogies in Brazilian (PT-BR) and European (PT-EU) Portuguese. In syntactic analogies, we have the following categories: adjective-to-adverb, opposite, comparative, superlative, present-participle, nationality-adjective, past-tense, plural, and plural-verbs. In semantic analogies, we have the following categories: capital-common-countries, capital-world, currency, city-in-state and family. In each category, we have examples of analogies with four words:

*adjective-to-adverb*

– *fantástico fantasticamente aparente aparentemente* **(syntactic)**
  fantastic fantastically apparent apparently

*capital-common-countries:*

– *Berlim Alemanha Lisboa Portugal* **(semantic)**
  Berlin Germany Lisbon Portugal

**Table 3** Accuracy values obtained in the intrinsic evaluation in syntactic and semantic analogies

| Embedding | PT-BR | | | PT-EU | | |
|---|---|---|---|---|---|---|
| | Syntactic | Semantic | All | Syntactic | Semantic | All |
| Word2Vec (word) | 49.4 | 42.5 | 45.9 | 49.5 | 38.9 | 44.3 |
| GloVe (word) | 34.7 | 36.7 | 35.7 | 34.9 | 34.0 | 34.4 |
| FastText (word) | 39.9 | 8.0 | 24.0 | 39.9 | 7.6 | 23.9 |
| MSSG (sense) | 23.0 | 6.6 | 14.9 | 23.0 | 6.3 | 14.7 |
| Sense2Vec (sense) | 52.4 | 42.6 | 47.6 | 52.6 | 39.5 | 46.2 |

*Algorithm*. The algorithm receives the first three words of the analogy and aims to predict the fourth. Thus, for instance considering the previous example, the algorithm would receive Berlin (a), Germany (b) and Lisbon (c) and should predict Portugal (d). Internally, the following algebraic operation is performed between vectors:

$$v(b) + v(c) - v(a) = v(d) \quad (1)$$

*Evaluation measure*. The measure used in this case is accuracy, which calculates the percentage of correctly labeled words in relation to the total amount of words in the dataset.

*Discussion of results*. Table 3 shows the general results of the accuracy obtained in the analogies.[11] Table 4 shows the results in each category of syntactic analogies for PT-BR. Table 5 shows the results for PT-EU. Table 6 shows the results in each category of semantic analogies for PT-BR. Table 7 shows the results for PT-EU. The Word2vec, GloVe and FastText were adopted as word embeddings baselines since they performed well in Hartmann et al. (2017) experiments. Note that the sense embeddings generated by our sense2vec model outperformed the baselines at the syntactic and semantic levels. MSSG, on the other hand, had the worst performance achieving the best accuracy of all models only in the comparative syntactic category.

In syntactic analogies, the sense embeddings generated by sense2vec outperformed the word embeddings generated by word2vec in opposite, nationality-adjective, past-tense, plural and plural-verbs as shown in Tables 4 and 5. An example of the adjective-to-adverb category is shown in Table 8. We can explain this type of success through an algebraic operation of vectors. When calculating v(*aparentemente* (apparently)) + v(*completo* (complete)) - v(*aparente* (apparent)) the resulting vector of word2vec is v(*incompleto* (incomplete)) when it should be v(*completamente* (completely)). The correct option appears as the second nearest neighbor.

[11] We have performed experiments varying the preprocessing steps (for example, using stems instead of surface forms) and the training corpus (only PT-BR or PT-EU) but the only experiment with a small gain was the one in which only PT-BR was used to train the Sense2Vec model. In this case, the accuracy in the semantic analogy was of 45.6 (a gain of 3 points) in PT-BR and of 41.3 (a gain of 1.8 points) in PT-EU.

**Table 4** Accuracy values obtained in each category in the syntactic analogies for PT-BR

| Category | PT-BR | | | | |
|---|---|---|---|---|---|
| | Word2Vec | GloVe | FastText | MSSG | Sense2Vec |
| Adjective-to-adverb | 14.4 | 5.1 | **23.4** | 5.9 | 11.0 |
| Opposite | 22.0 | 17.2 | 17.2 | 4.5 | **23.1** |
| Comparative | 66.7 | 60.0 | 56.7 | **73.3** | 60.0 |
| Superlative | 17.7 | 1.7 | **34.8** | 1.3 | 14.1 |
| Present-participle | **82.8** | 59.8 | 81.9 | 67.1 | 80.6 |
| Nationality-adjective | 69.2 | 49.0 | 10.6 | 2.6 | **73.7** |
| Past-tense | 53.3 | 41.2 | 58.4 | 38.7 | **62.9** |
| Plural | 44.0 | 33.3 | 38.9 | 15.0 | **49.4** |
| Plural-verbs | 47.8 | 30.5 | 50.2 | 35.5 | **51.6** |
| Total | 49.4 | 34.7 | 39.9 | 14.9 | **52.4** |

**Table 5** Accuracy values obtained in each category in the syntactic analogies for PT-EU

| Category | PT-EU | | | | |
|---|---|---|---|---|---|
| | Word2Vec | GloVe | FastText | MSSG | Sense2Vec |
| Adjective-to-adverb | 13.1 | 5.1 | **24.6** | 6.2 | 11.0 |
| Opposite | 22.0 | 17.2 | 17.2 | 4.5 | **23.1** |
| Comparative | 66.7 | 60.0 | 56.7 | **73.3** | 60.0 |
| Superlative | 17.7 | 1.7 | **34.8** | 1.3 | 14.1 |
| Present-participle | **83.1** | 58.5 | 82.4 | 66.8 | 80.9 |
| Nationality-adjective | 69.3 | 49.3 | 10.7 | 2.6 | **73.2** |
| Past-tense | 53.3 | 41.2 | 58.4 | 38.7 | **62.9** |
| Plural | 44.0 | 33.4 | 36.3 | 14.5 | **49.1** |
| Plural-verbs | 47.3 | 30.4 | 50.7 | 34.9 | **52.3** |
| Total | 49.5 | 34.9 | 39.9 | 23.0 | **52.6** |

Still analyzing the adjective-to-adverb syntactic category, which also contains analogies such as: *v(calmamente (quietly))* + *v(alegre (joyful))* - *v(calmo (calm))* = *v(alegremente (happily))* and *v(incerto (uncertain))* + *v(conveniente (convenient))* - *v(certo (right))* = *v(inconveniente (inconvenient))*, we concluded that FastText has the best accuracy (23.4% in PT-BR and 24.6% in PT-EU). One possible reason for FastText's best performance is that it prioritizes the morphology of words, generating vectors with shared weights between words of the same morphology (eg "travel" and "traveled"). The same effect occurs with the superlative syntactic category which contains analogies such as: *v(brilhantíssimo (brightest))* + *v(fácil (easy))* - *v(brilhante (bright))* = *v(facílimo (very easy))*.

**Table 6** Accuracy values obtained in each category in semantic analogies for PT-BR

| Category | PT-BR | | | | |
|---|---|---|---|---|---|
| | Word2Vec | GloVe | FastText | MSSG | Sense2Vec |
| Cap.-common-c. | **75.5** | 70.9 | 9.5 | 6.9 | 74.9 |
| Capital-world | 53.9 | 36.8 | 9.1 | 6.6 | **54.0** |
| Currency | 6.0 | 1.6 | 0.3 | 1.0 | **6.6** |
| City-in-state | 22.7 | **35.6** | 4.5 | 5.1 | 24.4 |
| Family | **64.7** | 56.3 | 27.3 | 23.8 | 60.5 |
| Total | 42.5 | 36.7 | 8.0 | 6.6 | **42.6** |

**Table 7** Accuracy values obtained in each category in semantic analogies for PT-EU

| Category | PT-EU | | | | |
|---|---|---|---|---|---|
| | Word2Vec | GloVe | FastText | MSSG | Sense2Vec |
| Cap.-common-c. | **78.4** | 68.8 | 11.3 | 8.2 | 75.8 |
| Capital-world | 50.0 | 34.7 | 8.4 | 6.2 | **51.0** |
| Currency | 5.7 | 1.4 | 0.4 | 1.0 | **6.9** |
| City-in-state | 17.7 | **31.2** | 3.6 | 3.9 | 19.0 |
| Family | **62.8** | 56.3 | 29.2 | 24.7 | 61.4 |
| Total | 38.9 | 34.0 | 7.6 | 6.3 | **39.5** |

**Table 8** Example of syntactic analogy predicted by word2vec and sense2vec

| | |
|---|---|
| Word2vec | Aparente aparentemente completo : completamente **(expected)** |
| | Aparente aparentemente completo : incompleto **(predicted)** |
| Sense2vec | Aparente\|ADJ aparentemente\|ADV completo\|ADJ : completamente\|ADV **(expected)** |
| | Aparente\|ADJ aparentemente\|ADV completo\|ADJ : completamente\|ADV **(predicted)** |

Another interesting analysis is in relation to the past-tense syntactic category where sense2vec has the best accuracy (62.9% in PT-BR and PT-EU) and FastText is the second one (58.4% in PT-BR and PT-EU). This category contains analogies such as: *v(dançou (danced)) + v(descrevendo (describing)) - v(dançando (dancing)) = v(descreveu (described))* and *v(foi (went)) + v(dizendo (saying)) - v(indo (going)) = v(disse (said))*. We noticed that FastText hits more cases where there is a morphology sharing (first example) and misses more cases where there this morphology sharing is not present (second example). We believe that these results are related to a certain addition of semantics in syntactic analogy (by using irregular verbs). Sense2vec, on the other hand, has a balanced performance in both cases, which means that it takes into account morphology and also understands semantics.

**Table 9** Example of semantic analogies predicted by word2vec and sense2vec

| | |
|---|---|
| word2vec | Arlington texas akron : kansas **(predicted)** ohio **(expected)** |
| sense2vec | Arlington\|N texas\|N akron\|N : ohio\|N **(predicted)(expected)** |
| word2vec | Bakersfield califórnia madison : pensilvânia **(predicted)** wisconsin **(expected)** |
| sense2vec | Bakersfield\|N califórnia\|N madison\|N : wisconsin\|N **(predicted)(expected)** |
| word2vec | Worcester massachusetts miami : seattle **(predicted)** flórida **(expected)** |
| sense2vec | Worcester\|N massachusetts\|N miami\|N : flórida\|N **(predicted)(expected)** |

The same effect occurs with the syntactic category plural-verbs, where there is also the use of irregular verbs (*v(vão (go)) + v(ver (to see)) − v(ir (to go)) = v(vêem (see))*).

In semantic analogies, the sense embeddings generated by sense2vec outperformed the word embeddings generated by word2vec in capital-world, currency and city-in-state. The Sense2vec receives corpus annotated with PoS tags, therefore, this additional information is used in the training and generation of sense embeddings. As much as the words of an analogy received the same PoS tag, the corresponding embeddings were generated based on the entire corpus. The training and generation of embeddings occur together with the generation of these vector spaces and PoS tags influence the distances of the embeddings. This ensure the quality of sense embeddings as a whole, that means, a better formation of vector spaces and its dimensions. An example of this case is presented in Table 9, where the PoS tag is always the same for all words (N, for noun) and sense2vec was able to predict correctly the analogy.

Still analyzing the city-in-state category, we observe that GloVe has the best accuracy (35.6% in PT-BR and 31.2% in PT-EU). This category contains analogies such as: *v(Colorado) + v(Orlando) − v(Denver) = v(Florida)*. We believe that this good performance is due to the fact that, according to Pennington et al. (2014), GloVe considers not only the local context as word2vec and sense2vec do, but also the global one.

So, the sense2vec's PoS tag functions as an extra feature in the training of sense embeddings, generating more accurate numerical vectors, allowing the correct result to be obtained. We also concluded that sense2vec handles morphological and semantic analogies well, absorbing some of the main features of FastText and GloVe at the same time.

#### *4.3.2 Extrinsic evaluation*

Sense representations are features that can be integrated in downstream tasks. The better the quality of these representations, the greater is the performance of these tasks. So we evaluated our sense embeddings in Semantic Textual Similarity (STS) and Word Sense Disambiguation (WSD) tasks. The deep neural language models were evaluated only in the STS task.

*4.3.2.1 Semantic textual similarity (STS)* This experiment is a task of semantic textual similarity between sentences where the use of sense embeddings and deep neural language models are evaluated. Word embeddings were chosen as baselines.

*Dataset*. ASSIN (Avaliação de Similaridade Semântica e Inferência Textual) was a workshop co-located with PROPOR-2016 which encompassed two shared-tasks regarding: (i) Semantic Textual Similarity and (ii) Recognizing Textual Entailment. We chose the first one to evaluate the models we implemented extrinsically in a semantic task. In ASSIN, the participants of the STS shared-task were asked to assign similarity values between 1 and 5 to pairs of sentences (gold score). The workshop made available the training and test sets for Brazilian (PT-BR) and European (PT-EU) Portuguese. This same dataset was used to fine-tune the BERT-Multilingual and BERT-Portuguese models.

*Algorithm*. The objective of this task is to predict, through a linear regression, the similarity score between two sentences. The model was trained in a dataset which contains sentence pairs with the gold score. The prediction occurs in the test set, which contains sentence pairs without the gold score. As we have this same test set with the gold score, it is possible to calculate Pearson's Correlation ($\rho$) and Mean Squared Error (MSE) between them. These results show how much the automatic prediction has approached the human prediction.

*Evaluation measures*. In this experiment, two evaluation measures were used: the Pearson's Correlation ($\rho$) and Mean Squared Error (MSE). $\rho$ measures the linear relationship between two datasets: one annotated by the participants (gold) and another which is output by the system. Like other correlation coefficients, this one

**Table 10** Extrinsic evaluation on STS task

| Embedding | Type | Size | PT-BR | | PT-EU | |
|---|---|---|---|---|---|---|
| | | | $\rho(\uparrow)$ | MSE ($\downarrow$) | $\rho(\uparrow)$ | MSE ($\downarrow$) |
| Word2Vec | Word | 300 | 0.56 | 0.52 | 0.55 | 0.83 |
| GloVe | Word | 300 | 0.46 | 0.60 | 0.46 | 0.93 |
| FastText | Word | 300 | 0.55 | 0.53 | 0.53 | 0.86 |
| MSSG | Sense | 300 | 0.40 | 0.64 | 0.35 | 1.04 |
| Sense2Vec | Sense | 300 | 0.56 | 0.52 | 0.53 | 0.87 |
| ELMo | lm | – | 0.61 | 0.47 | 0.62 | 0.74 |
| BERT-Multilingual | lm | – | 0.49 | 0.58 | 0.43 | 0.99 |
| BERT-Portuguese | lm | – | 0.55 | 0.53 | 0.51 | 0.91 |
| BERT-Multilingual† | lm | – | 0.78 | 0.33 | **0.82** | **0.44** |
| BERT-Portuguese† | lm | – | **0.80** | **0.33** | 0.82 | 0.47 |

Arrows indicate whether lower (↓) or higher (↑) is better varies between $-1$ and $+1$ with 0 meaning no correlation. Correlations of $-1$ or $+1$ mean an exact linear relationship. The MSE, in turn, measures the average of the squares of the errors, that is, the average squared difference between the estimated value and what was expected.

† Fine-tuning approach

*Discussion of results*. Table 10 shows the performance of sense embeddings and language models for both PT-BR and PT-EU test sets, through the Pearson's Correlation ($\rho$) and MSE. We apply the Significance Test for Pearson's Correlation Coefficient to ensure that the correlations did not occur by chance. The null hypothesis was rejected with 99% confidence for all correlations obtained, showing that these values are significant.

The results obtained by BERT-Multilingual (PT-EU) and BERT-Portuguese (PT-BR) fine-tuned models outperformed all other models and baselines, obtaining a higher correlation coefficient and a smaller error (results in bold). We attribute this performance to the use of the fine-tuning approach, which adjusted the model's weights in the STS task. According to Peters et al. (2019), the relative performance of fine-tuning vs. feature based depends on the similarity of the pre-training and the target tasks. This means that, as BERT is pre-trained in the next-sentence prediction objective, the performance in the STS task should be good, which is the case in our experiment.

The result obtained by ELMo model outperformed the sense embeddings models and the baselines. We can explain the good performance of ELMo through some features that it has. ELMo word representations are purely character-based, which allows the network to use morphological clues to form robust representations for out-of-vocabulary tokens unseen during training. Another point is that it incorporates everything it receives—characters, words, sentences or paragraphs.

The BERT-Multilingual and BERT-Portuguese feature based models, on the other hand, obtain a lower accuracy than expected for a deep neural language model. We attribute this performance to the use of the feature-based approach, which only uses the model's weights without improving them for the specific task.

Analyzing the performance of sense embeddings, we can notice that sense2vec performed better than MSSG and similar to word embeddings baselines, mainly word2vec.

In Table 11 we have two pairs of sentences (PT-BR) with the gold score (gold) and the scores generated by Word2vec, Sense2vec, ELMo, BERT-Multilingual (mBERT) and BERT-Portuguese (pBERT). In the first example, the score closest to gold was predicted by BERT-Multilingual and in the second, by ELMo.

*4.3.2.2 Word sense disambiguation (WSD)* Based on Pelevina et al. (2017) and Nóbrega (2013), this experiment is a lexical sample task (disambiguation of a word sample) where the precision of a disambiguation method using sense embeddings (MSSG) is evaluated. The most frequent sense method (MFS) was chosen as the baseline because it is widely used in the literature. Here it is also important to mention that only MSSG was evaluated as the sense embedding method because all words chosen for this lexical sample task were nouns, so sense2vec was not used as it would not be able to distinguish these senses.

*Dataset*. The CSTNews dataset (Cardoso et al. 2011) contains 50 collections of journalistic documents (PT-BR) with 72,148 words arranged in 140 documents grouped according to the topic addressed. According to Nóbrega (2013), the CSTNews dataset contains 466 nouns annotated with meanings, from the synsets of WordNet. Of this total, 361 have two or more meanings and of these, 196 have possibilities of above-average disambiguation found in the corpus, which is approximately 6 different senses. Thus, 77% of the words are ambiguous (with more than two meanings) and approximately 42% are highly ambiguous, with above average meanings.

*Algorithm*. To disambiguate a word in a context, we follow the algorithm proposed by Pelevina et al. (2017), which is based on similarity between sense and context. Given a target word $w$ and its context words $C = c_1, \ldots, c_k$, we first map $w$ to a set of its sense embeddings: $S = s_1, \ldots, s_n$. Instead of mapping the context words $C$ to sense embeddings, we map to their respective word embeddings (word2vec), avoiding a dimensionality problem. To improve WSD performance we also apply context filtering. Typically, only several words in context are relevant for sense disambiguation, like "chairs" and "kitchen" are for "table" in "They bought a table and chairs for kitchen." For each word $c_j$ in context $C = (c_1, \ldots, c_k)$ we calculate a score that quantifies how well it discriminates the senses:

$$max_i f(s_i, c_j) - min_i f(s_i, c_j) \tag{2}$$

**Table 11** Example of sentence pairs with similarity scores

| | |
|---|---|
| gold = **2.5** | *E já que meu período de Apple Music acabou, o jeito vai ser* |
| mBERT†= **2.5413015** | *voltar pro Spotify.* |
| mBERT = 2.8929672 | (And since my Apple Music period is over, the way will be |
| pBERT †= 2.5980363 | to go back to Spotify.) |
| pBERT = 3.0232697 | *Termina hoje o período de avaliação gratuita do Apple Music.* |
| ELMo = 2.568166 | (The free trial period for Apple Music ends today.) |
| sense2vec = 2.849287 | |
| word2vec = 2.9015808 | |
| gold = **1.75** | *O pedido foi feito pela Polícia Federal.* |
| ELMo = **1.891884** | (The request was made by the Federal Police.) |
| mBERT†= 2.5389488 | |
| mBERT = 2.5394192 | *Teori, no entanto, negou o pedido.* |
| pBERT†= 2.0468946 | (Teori, however, denied the request.) |
| pBERT = 2.2103405 | |
| sense2vec = 2.331877 | |
| word2vec = 2.413431 | |

† Fine-tuning approach

**Table 12** Lexical sample task

| Ambiguous word | Frequency | Senses | MFS Precision | Sense embedding Precision |
|---|---|---|---|---|
| obra | 13 | 2 | 71.59 | 65.81 |
| centro | 14 | 2 | 73.43 | **88.88** |
| estado | 10 | 2 | 36.00 | **80.00** |
| discurso | 9 | 2 | 30.86 | **63.88** |
| escola | 8 | 2 | 25.00 | **75.00** |
| presidente | 65 | 3 | 41.75 | 35.65 |
| pontos | 10 | 2 | 49.00 | **80.00** |
| Average | | | 46.80 | 69.88 |

MFS (Most Frequent Sense) and MSSG

where $s_i$ iterates over senses of the ambiguous word and $f$ is our disambiguation strategie: $sim(s_i, c_j)$. The $p$ most discriminative context words are used for disambiguation.

*Evaluation measure*. As in Pelevina et al. (2017), to calculate the performance of the disambiguation method, it was necessary to perform a mapping between the sense inventory of CSTNews and the sense embeddings generated by the MSSG method. Because of this mapping, it became possible to know when the method correctly classified the meaning of the word or not. Only the words that obtained a valid mapping were chosen for the lexical sample task, which is when their meanings in CSTNews have a corresponding sense embedding. There have been cases where the sense present in CSTNews did not appear as sense embedding and vice versa. As this task is a classification with unbalanced classes (more frequent and less frequent senses), the weighted precision was chosen. This measure get the precision for each class, and weight by the number of instances of each class. The result is the weighted precision rather than a simple arithmetic precision (which considers that all classes are equally important).

*Discussion of results*. Table 12 shows the weighted precision of the disambiguation method in same words, calculated using sense embeddings (MSSG) and MFS (baseline). The results show that in some ambiguous words the method using sense embeddings outperformed MFS (bold cases). The word “centro” (center) assumes two meanings in fourteen sentences. In two sentences with the sense of institution (center of predictions, NASA center) and in twelve with the sense of central area (midtown). This is an example where the largest class is the most frequent sense of the word in the corpus, a characteristic that benefits the MFS. For this reason, we use the metric precision weighted, which does not suffer from the imbalance of the classes. At the end of Table 12, we have a mean of the precisions and we found that sense embeddings outperformed the baseline in this task of lexical samples.

## 5 Conclusion and future work

In this paper we investigated the performance of sense representations for Portuguese (Brazilian and European) in some important NLP intrinsic and extrinsic tasks. To do so, we did the generation and evaluation of sense embeddings and compared them with traditional word embeddings and the state-of-the-art pre-trained language models in the feature and fine-tuning approaches.

For the intrinsic task of syntactic and semantic analogies, sense embeddings (sense2vec) outperformed the baselines of traditional word embeddings (word2vec, Glove, FastText). For the STS task, sense embeddings (sense2vec) performed similar to the baselines. For the lexical sample task of WSD, where the goal was to induce the correct meaning of an ambiguous word within a context, sense embeddings (MSSG) also obtained better precision when compared to the baseline of Most Frequent Sense. For the pre-trained language models (BERT and ELMo), BERT-Multilingual and BERT-Portuguese fine-tuned models outperformed ELMo, sense embeddings (Sense2vec and MSSG) and the baselines (Word2vec, GloVe and FastText) in the STS task.

These results show the effectiveness of these language resources for Portuguese tasks and allows a clear understanding of how unsupervised representations that capture meaning behave. Our generated sense embeddings models and the code used in all the experiments presented in this paper are available at https://github.com/LALIC-UFSCar/sense-embeddings.

As future work we intend to test other deep neural language models, as RoBERTa (Liu et al. 2019b) and Sentence-BERT (Reimers and Gurevych 2019). We also intend to do the evaluation of unsupervised sense representations in a multilingual way, to find out how much semantic representations generated for Portuguese are aligned with representations of other languages, and how to benefit from this alignment.

**Acknowledgements** This work was part of MMeaning Project supported by São Paulo Research Foundation (FAPESP—Regular Grant #2016/13002-0).